\documentclass{article}

\usepackage[preprint]{neurips_2025}
\usepackage{hyperref}
\usepackage{graphicx}
\usepackage{amsmath}
\usepackage{multirow}

\usepackage[utf8]{inputenc} 
\usepackage[T1]{fontenc}    
\usepackage{hyperref}       
\usepackage{url}            
\usepackage{booktabs}       
\usepackage{amsfonts}       
\usepackage{nicefrac}       
\usepackage{microtype}      
\usepackage{xcolor}         

\title{Less is More: Efficient Point Cloud Reconstruction via Multi-Head Decoders}

\author{%
Pedro Alonso$^{1,2}$ \quad Tianrui Li$^{1,2}$ \quad Chongshou Li$^{1,2}$\thanks{Corresponding Author} \\
$^1$ School of Computing and Artificial Intelligence, Southwest Jiaotong University\\
$^2$ Engineering Research Center of Sustainable Urban Intelligent Transportation \\
Ministry of Education, Chengdu, China\\
\texttt{\{palonso, trli, lics\}@swjtu.edu.cn}\\
}

\begin{document}

\maketitle

\begin{abstract}
  We challenge the common assumption that deeper decoder architectures always yield better performance in point cloud reconstruction. Our analysis reveals that, beyond a certain depth, increasing decoder complexity leads to overfitting and degraded generalization. Additionally, we propose a novel multi-head decoder architecture that exploits the inherent redundancy in point clouds by reconstructing complete shapes from multiple independent heads, each operating on a distinct subset of points. The final output is obtained by concatenating the predictions from all heads, enhancing both diversity and fidelity. Extensive experiments on ModelNet40 and ShapeNetPart demonstrate that our approach achieves consistent improvements across key metrics—including Chamfer Distance (CD), Hausdorff Distance (HD), Earth Mover's Distance (EMD), and F1-score—outperforming standard single-head baselines. Our findings highlight that output diversity and architectural design can be more critical than depth alone for effective and efficient point cloud reconstruction.
\end{abstract}

\section{Introduction} \label{section:Introduction}
Reconstructing point clouds—sparse, unordered 3D point sets captured by LiDAR or RGB-D sensors—into accurate, semantically meaningful shapes is essential for tasks like shape completion and generative modeling. Unlike images or voxels, point clouds lack grid structure and are inherently irregular, posing unique challenges for traditional deep learning models. Consequently, learning compact, informative latent representations becomes essential. Point cloud reconstruction both validates the quality of the encoding and enables a wide range of downstream tasks.

Pioneering architectures like PointNet \citep{Qi_2017_CVPR}, PointNet++ \citep{NIPS2017_d8bf84be}, and FoldingNet \citep{Yang_2018_CVPR} use shared point-wise feature extractors followed by a symmetric (e.g., max-pool) aggregation to achieve permutation invariance. Building on this foundation, several encoder-decoder models have achieved state-of-the-art results in reconstruction, classification, and segmentation. However, most still assume that: (1) deeper decoders improve reconstruction quality, and (2) a single decoder head suffices to reconstruct the entire point cloud.

In this work, we challenge these assumptions and investigate how decoder depth and design affect reconstruction quality. We demonstrate that deepening the decoder can lead to overfiting—memorizing redundant points rather than capturing shape semantics—and that traditional single-head decoders often prioritize point-level matching over accurate global structure. To overcome this, we propose a multi-head decoder design where each head reconstructs a disjoint subset of points, and their outputs are concatenated to form the complete shape. This design provides higher semantic meaning to each point, mitigates overfitting, and improves reconstruction quality.

We validate our approach on ModelNet40 \citep{Wu_2015_CVPR} and ShapeNetPart \citep{10.1145/2980179.2980238}, using three different backbones: Light Autoencoder (Light-AE), Deep Autoencoder (Deep-AE), and Point Transformer v3 (PTv3). Our experiments demonstrate that multi-head decoders consistently outperform standard single-head baselines across multiple metrics: Chamfer Distance, Hausdorff Distance, Earth Mover's Distance, and F1-score.

Our contributions are summarized as follows:

\textbullet~ We demonstrate that shallower decoder architectures can achieve superior structural fidelity in point cloud reconstruction, challenging the assumption that deeper decoders always perform better.

\textbullet~ We propose a novel multi-head decoder architecture that reconstructs complete shapes from multiple independent heads, in which each head operates on a subset of points.

\textbullet~ We conduct extensive experiments across multiple architectures and datasets, demonstrating that our multi-head approach consistently outperforms standard single-head decoders across a range of metrics, including CD, EMD, HD, and F1-score.

\section{Related work} \label{section:Related_work}
\subsection{Point cloud reconstruction}
Point cloud reconstruction is a fundamental task in 3D vision, with applications in downstream tasks such as shape completion, object classification, and semantic segmentation. Over time, a wide range of methods have been proposed to address this challenge, ranging from early autoencoder-based methods to more recent diffusion models.

\textbf{Autoencoder-based methods:}
Autoencoder-based methods laid the foundation for point cloud reconstruction. FoldingNet \citep{Yang_2018_CVPR} introduced a decoder that "folds" a 2D grid into a 3D surface, learning meaningful deformations. AtlasNet \citep{vakalopoulou:hal-01958236} extended this idea by learning multiple small patches (atlases), each decoded separately and then merged to form the full 3D shape. PCN (Point Completion Network; \citep{yuan2018pcn}) uses a two-stage decoder that first generates a coarse point cloud from the latent space and then refines it via a local folding network. Other approaches, such as TopNet \citep{Tchapmi_2019_CVPR} and \citep{liu2020morphing}, introduced hierarchical structures and multistage generation to improve reconstruction fidelity.

\textbf{GAN-based approaches:}
GAN-based approaches, such as 3D-GAN \citep{NIPS2016_44f683a8} and PC-GAN \citep{li2018pointcloudgan} introduced adversarial training for generating 3D shapes. 3D-GAN trains a volumetric GAN on 3D occupancy grids, sampling novel shapes from a learned latent distribution, while PC-GAN operates directly on raw point clouds using graph-based discriminators and generators. By contrast, SnowflakeNet \citep{Xiang_2021_ICCV} tackles point cloud completion through a coarse-to-fine, tree-structured, and self-attentive feature expansion, achieving high-resolution outputs.

\textbf{Transformer-based models:}
Transformer-based models have recently been applied to shape completion and generation. Point-BERT \citep{yu2022point} leverages masked point modeling—randomly hiding subsets of points and training a transformer model to reconstruct them—to learn rich 3D shape representations. ShapeFormer \citep{Yan_2022_CVPR} builds a hierarchical transformer that processes point clouds at multiple scales, capturing both local and global features for shape understanding. PoinTr \citep{Yu_2021_ICCV} treats point cloud completion as a sequence prediction task, using a transformer decoder to autoregressively generate the missing points.

\textbf{Diffusion models:}
Diffusion models have also been applied to point cloud reconstruction and generation. Point-E \citep{nichol2022point} uses a series of diffusion models to generate 3D shapes from text or latent vectors. Similarly, \citep{Luo_2021_CVPR} has shown promising results in denoising-based reconstruction pipelines.


\subsection{Multi-head Architectures}
While most point cloud reconstruction methods employ single-head decoders to reconstruct the entire shape from a latent representation, recent works have shown that multi-head decoder architectures can enhance reconstruction performance.

For instance, \citep{liu2020morphing} performs a two stage, parallel decoder reconstruction, where one branch captures the main object geometry and the other progressively upsamples additional points to produce the final shape. This split into coarse and fine decoding demonstrates how parallel decoding can enhance reconstruction. Similarly, tree-GAN \citep{Shu_2019_ICCV} employs a tree-structured decoder for hierarchical point generation, introducing a branching mechanism to capture both global and local shape features.

AtlasNet \citep{vakalopoulou:hal-01958236} can also be seen as a multi-branch decoder, where multiple MLPs independently generate local surface patches, which are then merged into the final shape.

Ensemble-based and multi-stage strategies are also common. For instance, PCN \citep{yuan2018pcn} employs a coarse-to-fine decoder with intermediate supervision, while GRNet \citep{xie2020grnet} merges grid-based partial predictions into a unified point set. Although not explicitly multi-head, these approaches reflect the underlying intuition that decomposing the output space can enhance generalization.


\section{Method}
\begin{figure}
  \centering
  \includegraphics[width=1\textwidth]{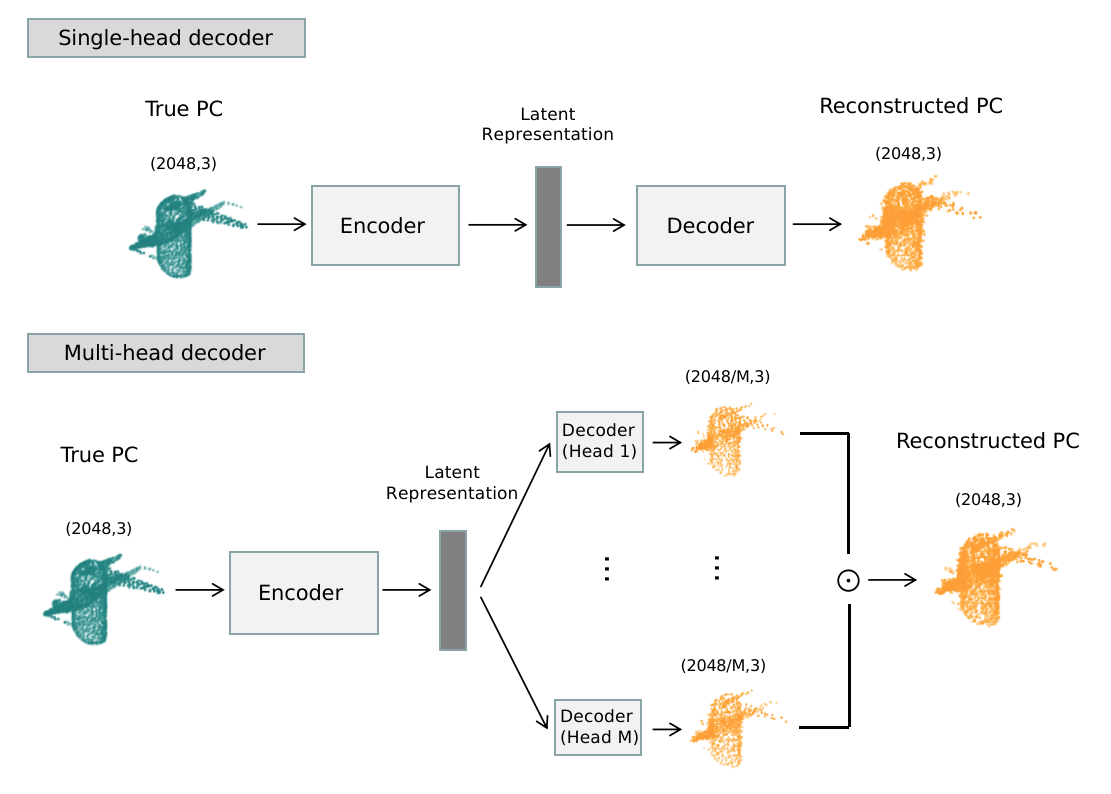}
  \caption{Comparison of single-head and multi-head decoder architectures. The top row shows the standard encoder–decoder pipeline, where a single-head decoder reconstructs the entire point cloud from the latent representation. The bottom row illustrates our multi-head design, in which M parallel decoders each generate a subset of points, which are then concatenated to form the final reconstructed point cloud.}
  \label{fig:plot_1}
  \vspace{-1.5em}
\end{figure}

\subsection{Backbone architectures}
\label{sec:baselines}
We evaluate our method on three backbone architectures: a light autoencoder (Light-AE), a PointNet-like deep autoencoder (Deep-AE), and Point Transformer v3 (PTv3).

\textbf{Light-AE encoder: } A lightweight encoder composed of three 1D convolutional layers: $3\to64$ , $64\to128$ , and $128\to128$, each followed by batch normalization and ReLU, except the last layer which omits ReLU. A global max-pooling layer reduces the output to a 128-dimensional latent vector.

\textbf{Deep-AE encoder: } A PointNet-inspired encoder with five 1D convolutional blocks: $3\to64$ , $64\to64$ , $64\to64$ , $64\to128$ , and $128\to1024$, each followed by batch normalization and ReLU (except the last layer, which omits ReLU). A global max-pooling operation reduces the output to a 1024-dimensional latent vector.

\textbf{PTv3 encoder: } We use Point Transformer v3 (PTv3; \citep{wu2024point}) with only $(x, y, z)$ coordinates as input and \textit{FlashAttention} disabled. The network applies an initial embedding followed by five stages of transformer blocks to produce per-point features. A global max-pooling operation on the final features produces a 64-dimensional vector. We add an extra MLP layer that projects the 64-dimensional vector into a final 512‑dimensional latent vector.

\begin{table}
  \caption{Decoder configurations by depth (1 to 5 layers) for each backbone: Light-AE, Deep-AE, and PTv3. We report the number of parameters for single-head and two-head configurations.}
  \label{tab:decoder_config}
  \centering
  \begin{tabular}{clcc}
    \toprule
    \multirow{2}{*}{\textbf{Layers}} & \multirow{2}{*}{\textbf{\hspace{4.2em}Light-AE decoder}} & \multicolumn{2}{c}{\textbf{Parameters (M)}} \\
    \cmidrule(lr){3-4}
    & & \textbf{Single-head} & \textbf{Two-head} \\
    \midrule
    1 & $ 128 \to 2048\times3$ & 0.79 & 0.79\\
    2 & $ 128 \to 256 \to 2048\times3$ & 1.61 & 1.65\\
    3 & $ 128 \to 256 \to 512 \to 2048\times3$ & 3.31 & 3.48\\
    4 & $ 128 \to 256 \to 512 \to 1024 \to 2048\times3$ & 6.99 & 7.68\\
    5 & $ 128 \to 256 \to 512 \to 1024 \to 1024 \to 2048\times3$ & 8.04 & 9.78\\
    \midrule
    \multicolumn{2}{c}{\textbf{Deep-AE decoder}} \\
    \midrule
    1 & $ 1024 \to 2048\times3$ & 6.30 & 6.30\\
    2 & $ 1024 \to 512 \to 2048\times3$ & 3.67 & 4.20\\
    3 & $ 1024 \to 512 \to 1024 \to 2048\times3$ & 7.35 & 8.40\\
    4 & $ 1024 \to 512 \to 1024 \to 1024 \to 2048\times3$ & 8.40 & 10.50\\
    5 & $ 1024 \to 512 \to 1024 \to 1024 \to 1024 \to 2048\times3$ & 9.45 & 12.60\\
    \midrule
    \multicolumn{2}{c}{\textbf{PTv3 decoder}} \\
    \midrule
    1 & $ 512 \to 2048\times3$ & 3.15 & 3.15\\
    2 & $ 512 \to 256 \to 2048\times3$ & 1.71 & 1.84\\
    3 & $ 512 \to 256 \to 512 \to 2048\times3$ & 3.41 & 3.68\\
    4 & $ 512 \to 256 \to 512 \to 1024 \to 2048\times3$ & 7.09 & 7.87\\
    5 & $ 512 \to 256 \to 512 \to 1024 \to 1024 \to 2048\times3$ & 8.14 & 9.97\\
    \bottomrule
  \end{tabular}
  \vspace{-1.5em}
\end{table}

\subsection{Metrics} \label{section:metrics}
We evaluate point cloud reconstruction quality using four standard metrics:

\textbf{Chamfer Distance (CD): }
Measures the average squared distance between nearest neighbors in two point clouds P and Q as,

\begin{equation}
    \text{CD}(P, Q) = \frac{1}{N_P} \sum_{p \in P} \min_{q \in Q} \|p - q\|^2 + \frac{1}{N_Q} \sum_{q \in Q} \min_{p \in P} \|q - p\|^2,
    \label{eq:cd}
\end{equation}

where $N_P$ and $N_Q$ are the number of points in P and Q, respectively.

\textbf{Earth Mover's Distance (EMD): } Quantifies the minimum cost required to transform one point cloud into another. It is calculated as,

\begin{equation}
    \text{EMD}(P, Q) = \min_{\phi: P \to Q} \frac{1}{|P|} \sum_{p \in P} \|p - \phi(p)\|_2^2,
    \label{eq:emd}
\end{equation}
where $\phi$ denotes bijection.

\textbf{Hausdorff Distance (HD): } Captures the worse-case deviation between two point clouds as,

\begin{equation}
    \text{HD}(P, Q) = \max \left\{ \sup_{p \in P} \inf_{q \in Q} \|p - q\|_2,\; \sup_{q \in Q} \inf_{p \in P} \|q - p\|_2 \right\}.
\end{equation}

\textbf{F1 Score at $1\%$ ($F1@1\%$): } Measures geometric similarity by combining precision and recall as,

\begin{equation}
    \text{F1} = \frac{2 \cdot \text{Precision} \cdot \text{Recall}}{\text{Precision} + \text{Recall}}.
\end{equation}

A point is considered a true positive (TP) if it lies within 1\% of the bounding box diagonal from its nearest neighbor in the other point cloud; otherwise, it is considered a false positive (FP).

\subsection{Multi-head decoder}
Standard point cloud decoder designs typically reconstruct the entire point cloud from a single-head decoder. However, these methods face two key limitations: (1) point cloud sparsity causes the model to overfit to redundant or overrepresented areas, and (2) these decoders tend to prioritize point-level accuracy at the expense of global shape semantics.

To address these issues, we propose a multi-head decoder design (Figure~\ref{fig:plot_1}). Our approach decomposes the reconstruction task across M independent heads that share the same latent vector as input. Each head is responsible for a disjoint subset of points (K/M, where K is the number of input points). The outputs from all heads are then concatenated to form the final point cloud of K points. This design offers two key advantages:

\textbf{Enhanced semantic representation: } By reducing each head’s output size, every point is encouraged to encode richer semantic information, as each head independently recovers the entire shape. This setup promotes better coverage of underrepresented regions and reduces redundancy.

\textbf{Training efficiency: } Thanks to the Chamfer Distance's ability to compare point clouds with differing number of points, the loss for multi-head models can be computed efficiently as the average CD between each head's output and the ground truth.

For a decoder of M-heads, the CD is calculated as:

\begin{equation}
    \label{eq:cd_multihead}
    CD(P, Q_1,...,Q_M) = \frac{1}{M} \sum_{i=1}^{M} \left[\frac{1}{N_P} \sum_{p \in P} \min_{q \in Q_i} \|p - q\|^2 + \frac{1}{N_{Q_i}} \sum_{q \in Q_i} \min_{p \in P} \|q - p\|^2\right],
\end{equation}

where P is the true point cloud and $Q_1, ..., Q_M$ are the predicted point clouds of each head.

\begin{table}
  \caption{Performance comparison on ModelNet40 for three models; Light-AE, Deep-AE, and PTv3. For each model, we evaluate single-head and multi-head decoders across varying depths. All models were trained for 100 epochs. Multi-head performance changes (green: improvement, red: decline) are relative to the single-head counterparts. The best score for each metric is highlighted in bold.}
  \label{tab:results_modelnet40}
  \centering
  \begin{tabular}{lccccc}
    \toprule
    \multicolumn{6}{c}{\textbf{Light-AE}}\\
    \midrule
    \textbf{Model} & \textbf{Layers} & \textbf{$10^3 \times$CD} & \textbf{EMD} & \textbf{$10^2 \times$HD} & \textbf{F1@1$\%$ score} \\
    \midrule
     & 1 & 3.39 & 154.74 & 16.97 & 24.84 \\
     & 2 & 3.52 & 150.88 & 16.92 & 23.96 \\
    Single & 3 & 3.29 & 138.40 & 15.79 & 26.03 \\
     & 4 & 3.38 & 145.12 & 15.61 & 26.94 \\
     & 5 & 3.48 & 166.64 & 16.08 & 26.76 \\
    \midrule
     &  1 & 3.42 (\textcolor{red}{$\downarrow$0.03}) & 128.37 (\textcolor{green}{$\downarrow$26.37}) & 17.07 (\textcolor{red}{$\uparrow$0.10}) & 25.37 (\textcolor{green}{$\uparrow$0.53}) \\
     &  2 & 3.37 (\textcolor{green}{$\uparrow$0.15}) & 110.20 (\textcolor{green}{$\downarrow$40.68}) & 16.73 (\textcolor{green}{$\downarrow$0.19}) & 25.16 (\textcolor{green}{$\uparrow$1.20}) \\
    Multi-head &  3 & \textbf{3.18} (\textcolor{green}{$\uparrow$0.11}) & \textbf{103.92} (\textcolor{green}{$\downarrow$34.48}) & 15.77 (\textcolor{green}{$\downarrow$0.02}) & 26.83 (\textcolor{green}{$\uparrow$0.80}) \\
     &  4 & 3.24 (\textcolor{green}{$\uparrow$0.14}) & 111.50 (\textcolor{green}{$\downarrow$33.62}) & \textbf{15.42} (\textcolor{green}{$\downarrow$0.19}) & \textbf{27.45} (\textcolor{green}{$\uparrow$0.51}) \\
     &  5 & 3.36 (\textcolor{green}{$\uparrow$0.12}) & 113.09 (\textcolor{green}{$\downarrow$53.55}) & 15.61 (\textcolor{green}{$\downarrow$0.47}) & 27.37 (\textcolor{green}{$\uparrow$0.61}) \\
    \midrule
    \multicolumn{6}{c}{\textbf{Deep-AE}}\\
    \midrule
     & 1 & 3.19 & 143.18 & 15.37 & 25.43 \\
     & 2 & 3.41 & 119.96 & 15.74 & 23.62 \\
    Single & 3 & 3.14 & 118.16 & 15.11 & 26.33 \\
     & 4 & 3.05 & 123.62 & 15.03 & 27.30 \\
     & 5 & 3.18 & 125.20 & 15.63 & 26.76 \\
    \midrule
     & 1 & 3.08 (\textcolor{green}{$\downarrow$0.11}) & 122.29 (\textcolor{green}{$\downarrow$20.89}) & 15.12 (\textcolor{green}{$\downarrow$0.25}) & 26.49 (\textcolor{green}{$\uparrow$1.06}) \\
     & 2 & 3.41 (\textcolor{black}{$=$}) & 107.78 (\textcolor{green}{$\downarrow$12.18}) & 15.95 (\textcolor{red}{$\uparrow$0.21}) & 22.89 (\textcolor{red}{$\downarrow$0.73}) \\
    Multi-head &3 & 2.99 (\textcolor{green}{$\downarrow$0.15}) & 98.00 (\textcolor{green}{$\downarrow$20.16}) & \textbf{14.64} (\textcolor{green}{$\downarrow$0.47}) & 26.56 (\textcolor{green}{$\uparrow$0.23}) \\
     & 4 & \textbf{2.96} (\textcolor{green}{$\downarrow$0.09}) & \textbf{91.47} (\textcolor{green}{$\downarrow$32.15}) & 14.71 (\textcolor{green}{$\downarrow$0.32}) & \textbf{27.67} (\textcolor{green}{$\uparrow$0.37}) \\
     & 5 & 3.08 (\textcolor{green}{$\downarrow$0.10}) & 98.40 (\textcolor{green}{$\downarrow$26.80}) & 15.18 (\textcolor{green}{$\downarrow$0.45}) & 27.10 (\textcolor{green}{$\uparrow$0.34}) \\
    \midrule
    \multicolumn{6}{c}{\textbf{Deep-AE}}\\
    \midrule
     & 1 & \textbf{2.72} & 113.50 & 16.19 & 29.11 \\
     & 2 & 2.89 & 128.68 & 16.20 & 28.06 \\
    Single & 3 & 2.85 & 133.19 & 15.61 & 28.99 \\
     & 4 & 2.95 & 155.89 & 15.22 & 29.14 \\
     & 5 & 3.12 & 183.84 & 15.61 & 28.90 \\
    \midrule
     & 1 & 2.76 (\textcolor{red}{$\uparrow$0.04}) & \textbf{94.73} (\textcolor{green}{$\downarrow$18.77}) & 16.07 (\textcolor{green}{$\downarrow$0.12}) & 28.74 (\textcolor{red}{$\downarrow$0.37}) \\
     & 2 & 2.77 (\textcolor{green}{$\downarrow$0.12}) & 107.79 (\textcolor{green}{$\downarrow$20.89}) & 15.60 (\textcolor{green}{$\downarrow$0.60}) & 28.61 (\textcolor{green}{$\uparrow$0.55}) \\
    Multi-head & 3 & 2.77 (\textcolor{green}{$\downarrow$0.08}) & 107.70 (\textcolor{green}{$\downarrow$25.49}) & 15.12 (\textcolor{green}{$\downarrow$0.49}) & 29.03 (\textcolor{green}{$\uparrow$0.04}) \\
     & 4 & 2.86 (\textcolor{green}{$\downarrow$0.09}) & 103.40 (\textcolor{green}{$\downarrow$52.49}) & \textbf{14.98} (\textcolor{green}{$\downarrow$0.24}) & \textbf{29.66} (\textcolor{green}{$\uparrow$0.52}) \\
     & 5 & 3.00 (\textcolor{green}{$\downarrow$0.12}) & 111.18 (\textcolor{green}{$\downarrow$72.66}) & 15.17 (\textcolor{green}{$\downarrow$0.44}) & 29.10 (\textcolor{green}{$\uparrow$0.20}) \\
    \bottomrule
  \end{tabular}
  \vspace{-2em}
\end{table}

\section{Experiments}
In this section, we describe the experimental setup, including the datasets used, backbone architectures, and training and evaluation details. We then present extensive experiments that analyze the impact of decoder depth and the benefits of multi-head decoders, followed by a discussion of our findings.
\subsection{Experimental setup}
\textbf{Datasets: }
We evaluate our models on two standard benchmarks: ModelNet40 and ShapeNetPart.

\textit{ModelNet40: }ModelNet40 \citep{Wu_2015_CVPR} is a large-scale dataset for 3D shape understanding, consisting of 12,311 pre-aligned shapes across 40 categories. It is widely used for tasks such as object classification, segmentation, and point cloud reconstruction. The point clouds consist of 2048 points. The dataset is split into 9,840 point clouds for training ($80\%$) and 2468 for testing ($20\%$).

\textit{ShapeNetPart: }ShapeNetPart \citep{10.1145/2980179.2980238} is derived from the larger ShapeNet dataset \citep{chang2015shapenetinformationrich3dmodel}. It includes 16 object categories and annotated per-point part labels (not used in this work, we use only the spatial coordinates $(x, y, z)$). The data is split into 12137 shapes for training ($70\%$), 1,870 for validation ($10\%$), and 2,874 for testing ($20\%$). Each point cloud is normalized and consists of 2048 points.

\textbf{Backbones:}
We evaluate three encoder backbones—Light-AE, Deep-AE, and PTv3 (see Section~\ref{sec:baselines})—each paired with decoders of five different depths (1–5 layers) and tested in both single-head and multi-head configurations. In the experiments presented in this work we use two heads for the multi-head models.

All decoder layers are implemented as MLPs with ReLU activations, except for the final layer which does not include ReLU. Batch normalization is excluded from Light-AE's decoder, since it degrades performance at lower depths. In contrast, Deep-AE benefits from batch normalization, which we apply after each layer except the last. We do not include batch normalization in PTv3's decoder.

Decoder configurations are presented in Table~\ref{tab:decoder_config}. For two-head decoders, each head follows the same layer sequence as the single-head variant, except for the final layer, which is reduced to output $1024\times3$ points. The outputs of both heads are then concatenated to form the full $2048\times3$ point cloud.

\textbf{Training details: }
All models are trained using the Chamfer distance as the loss function, following Eq.~\ref{eq:cd_multihead} with $M=1$ for single-head models and $M=2$ two head decoders. We use the Adam optimizer \citep{kingma2014adam} with a learning rate of $5\cdot10^{-4}$ for Light-AE and Deep-AE, and $1\cdot10^{-4}$ for PTv3. The batch size is set to 32 for Light-AE and Deep-AE, and 8 for PTv3. All models are trained for 100 epochs, and we report results from the epoch with the best performance on the test set.

\textbf{Evaluation metrics: }
We evaluate reconstruction performance using four metrics: Chamfer Distance (CD), Hausdorff Distance (HD), Earth Mover's Distance (EMD), and F1@1$\%$ score, following the definitions in Section~\ref{section:metrics}.

\subsection{Main results}
Tables~\ref{tab:results_modelnet40} and~\ref{tab:results_shapenetpart} present our experimental results across all backbones and decoder configurations on ModelNet40 and ShapeNetPart, respectively. Overall, we observe that multi-head decoder configurations consistently outperform their single-head counterparts across all evaluation metrics. The average improvements of multi-head models over single-head models are as follows:

\textbf{ModelNet40: } $\textbf{2.73\%}$ (CD)\hspace{0.2em},\hspace{0.3em} $\textbf{22.57\%}$ (EMD)\hspace{0.2em},\hspace{0.3em} $\textbf{1.68\%}$ (HD)\hspace{0.2em},\hspace{0.3em} $\textbf{1.48\%}$ (F1-score).

\textbf{ShapeNetPart: } $\textbf{3.37\%}$ (CD)\hspace{0.2em},\hspace{0.3em} $\textbf{17.14\%}$ (EMD)\hspace{0.2em},\hspace{0.3em} $\textbf{2.42\%}$ (HD)\hspace{0.2em},\hspace{0.3em} $\textbf{1.23\%}$ (F1-score).

These improvements are consistent across different models and datasets, demonstrating that multi-head decoder designs enhance reconstruction in an architecture-agnostic and reliable manner.

\begin{figure}
  \centering
  \includegraphics[width=1\textwidth]{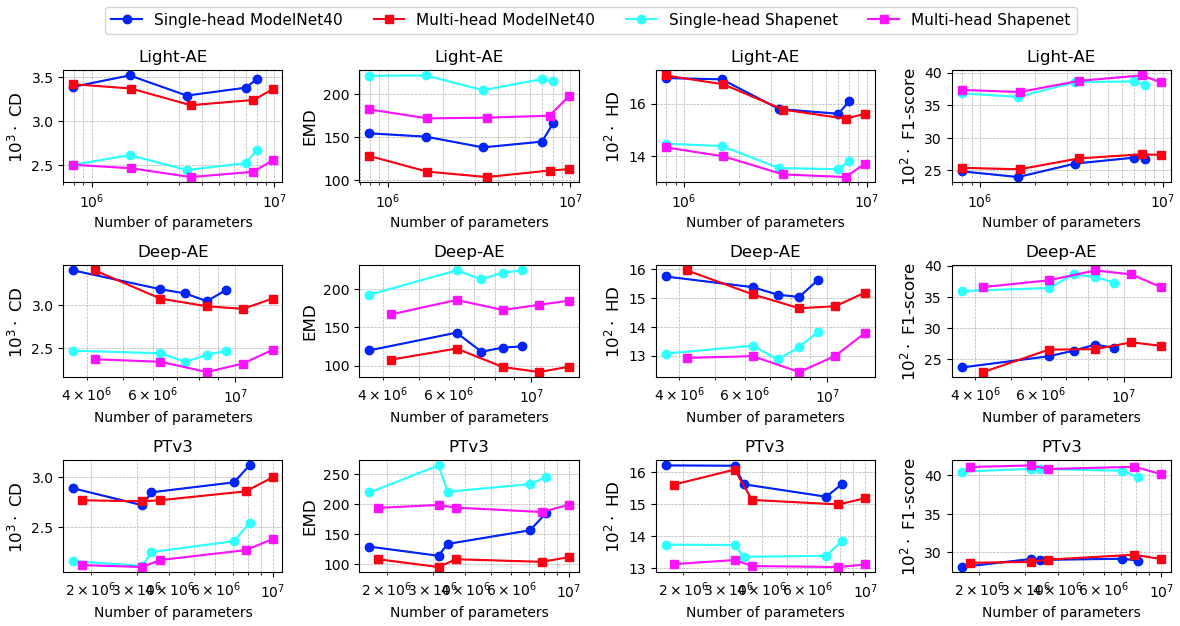}
  \caption{Reconstruction metrics as a function of decoder size for three backbone models; Light-AE (top row), Deep-AE (middle row), and PTv3 (bottom row). Within each panel, the x-axis shows the decoder depth, comparing single-head and multi-head configurations across two datasets: ModelNet40 and ShapeNetPart.}
  \label{fig:plot_metrics_vs_nparams}
  \vspace{-1em}
\end{figure}

Figure~\ref{fig:plot_metrics_vs_nparams} shows reconstruction performance as a function of decoder parameter count across all three backbone models and both datasets. Notably, single-head and multi-head configurations follow nearly identical trends within each architecture, further confirming that the benefits of multi-head designs are independent of model size or complexity.






\begin{table}
  \caption{Performance comparison on ShapeNetPart for three models; Light-AE, Deep-AE, and PTv3. For each model, we evaluate single-head and multi-head decoders across varying depths. All models were trained for 100 epochs. Multi-head performance changes (green: improvement, red: decline) are relative to the single-head counterparts. The best score for each metric is highlighted in bold.}
  \label{tab:results_shapenetpart}
  \centering
  \begin{tabular}{lccccc}
    \toprule
    \multicolumn{6}{c}{\textbf{Light-AE}}\\
    \midrule
    \textbf{Model}    & \textbf{Layers} & \textbf{$10^3\times$CD} 
                      & \textbf{EMD}    & \textbf{$10^2\times$HD} 
                      & \textbf{F1@1\%}\\
    \midrule
         & 1 & 2.50  & 221.60 & 14.47 & 36.82 \\
         & 2 & 2.61  & 222.07 & 14.38 & 36.27 \\
    Single & 3 & 2.44  & 204.88 & 13.54 & 38.54 \\
         & 4 & 2.52  & 217.72 & 13.50 & 38.65 \\
         & 5 & 2.67  & 215.04 & 13.81 & 38.06 \\
    \midrule
         & 1 & 2.50 ({\color{black}$=$})               & 182.53 (\textcolor{green}{$\downarrow$39.07}) & 14.33 (\textcolor{green}{$\downarrow$0.14}) & 37.34 (\textcolor{green}{$\uparrow$0.52}) \\
         & 2 & 2.46 (\textcolor{green}{$\uparrow$0.15}) & \textbf{172.15} (\textcolor{green}{$\downarrow$49.92}) & 13.99 (\textcolor{red}{$\uparrow$0.39}) & 37.02 (\textcolor{green}{$\uparrow$0.75}) \\
    Multi-head & 3 & \textbf{2.36} (\textcolor{green}{$\uparrow$0.08}) & 172.93 (\textcolor{green}{$\downarrow$31.95}) & 13.30 (\textcolor{green}{$\downarrow$0.24}) & 38.74 (\textcolor{green}{$\uparrow$0.20}) \\
         & 4 & 2.42 (\textcolor{green}{$\uparrow$0.10}) & 175.23 (\textcolor{green}{$\downarrow$42.49}) & \textbf{13.20} (\textcolor{green}{$\downarrow$0.30}) & \textbf{39.57} (\textcolor{green}{$\uparrow$0.92}) \\
         & 5 & 2.55 (\textcolor{green}{$\uparrow$0.12}) & 197.61 (\textcolor{green}{$\downarrow$17.43}) & 13.71 (\textcolor{green}{$\downarrow$0.10}) & 38.50 (\textcolor{green}{$\uparrow$0.44}) \\
    \midrule
    \multicolumn{6}{c}{\textbf{Deep-AE}}\\
    \midrule
         & 1 & 2.44  & 224.95 & 13.34 & 36.39 \\
         & 2 & 2.47  & 192.99 & 13.07 & 35.91 \\
    Single & 3 & 2.34  & 213.43 & 12.87 & 38.57 \\
         & 4 & 2.42  & 222.00 & 13.29 & 38.16 \\
         & 5 & 2.47  & 224.92 & 13.83 & 37.24 \\
    \midrule
         & 1 & 2.34 (\textcolor{green}{$\downarrow$0.10}) & 186.11 (\textcolor{green}{$\downarrow$38.84}) & 12.98 (\textcolor{green}{$\downarrow$0.36}) & 37.63 (\textcolor{green}{$\uparrow$1.24}) \\
         & 2 & 2.37 (\textcolor{green}{$\downarrow$0.10}) & \textbf{167.18} (\textcolor{green}{$\downarrow$25.81}) & 12.92 (\textcolor{green}{$\downarrow$0.15}) & 36.55 (\textcolor{green}{$\uparrow$0.64}) \\
    Multi-head & 3 & \textbf{2.22} (\textcolor{green}{$\downarrow$0.12}) & 172.99 (\textcolor{green}{$\downarrow$40.44}) & \textbf{12.43} (\textcolor{green}{$\downarrow$0.44}) & \textbf{39.22} (\textcolor{green}{$\uparrow$0.65}) \\
         & 4 & 2.32 (\textcolor{green}{$\downarrow$0.10}) & 179.84 (\textcolor{green}{$\downarrow$42.16}) & 12.99 (\textcolor{green}{$\downarrow$0.30}) & 38.57 (\textcolor{green}{$\uparrow$0.41}) \\
         & 5 & 2.48 (\textcolor{red}{$\uparrow$0.01})  & 185.08 (\textcolor{green}{$\downarrow$39.84}) & 13.78 (\textcolor{green}{$\downarrow$0.05}) & 36.54 (\textcolor{red}{$\downarrow$0.70}) \\
    \midrule
    \multicolumn{6}{c}{\textbf{PTv3}}\\
    \midrule
         & 1 & \textbf{2.11} & 263.86 & 13.71 & 40.84 \\
         & 2 & 2.16  & 218.84 & 13.72 & 40.48 \\
    Single & 3 & 2.25  & 220.08 & 13.34 & 40.84 \\
         & 4 & 2.36  & 232.67 & 13.37 & 40.59 \\
         & 5 & 2.54  & 244.46 & 13.83 & 39.74 \\
    \midrule
         & 1 & 2.10 (\textcolor{green}{$\downarrow$0.01}) & 198.07 (\textcolor{green}{$\downarrow$65.79}) & 13.24 (\textcolor{green}{$\downarrow$0.47}) & \textbf{41.29} (\textcolor{green}{$\uparrow$0.45}) \\
         & 2 & 2.12 (\textcolor{green}{$\downarrow$0.04}) & 193.10 (\textcolor{green}{$\downarrow$25.74}) & 13.11 (\textcolor{green}{$\downarrow$0.61}) & 41.08 (\textcolor{green}{$\uparrow$0.60}) \\
    Multi-head & 3 & 2.17 (\textcolor{green}{$\downarrow$0.08}) & 193.57 (\textcolor{green}{$\downarrow$26.51}) & 13.05 (\textcolor{green}{$\downarrow$0.29}) & 40.82 (\textcolor{red}{$\downarrow$0.02}) \\
         & 4 & 2.27 (\textcolor{green}{$\downarrow$0.09}) & \textbf{186.35} (\textcolor{green}{$\downarrow$46.32}) & \textbf{13.02} (\textcolor{green}{$\downarrow$0.35}) & 41.11 (\textcolor{green}{$\uparrow$0.52}) \\
         & 5 & 2.38 (\textcolor{green}{$\downarrow$0.16}) & 198.35 (\textcolor{green}{$\downarrow$46.11}) & 13.10 (\textcolor{green}{$\downarrow$0.73}) & 40.17 (\textcolor{green}{$\uparrow$0.43}) \\
    \bottomrule
  \end{tabular}
\end{table}






\subsubsection{Discussions}
\textbf{Impact of decoder depth vs parameter count: } Figure~\ref{fig:plot_metrics_vs_nparams} reveals a consistent trend: reconstruction performance aligns more closely with decoder depth than with parameter count. This suggests that architectural depth—rather than raw capacity—plays a more significant role in reconstruction quality. While increasing the number of parameters (e.g., via wider layers) enhances representational capacity, it does not change the fundamental structure of the decoder. In contrast, adding more layers introduces additional abstraction steps, which can have a greater—positive or negative— impact the model's ability to capture shape semantics.

\textbf{EMD and richer semantic representations: }Earth Mover’s Distance (EMD) shows the most significant improvement among all metrics when using multi-head decoders. Interestingly, EMD values remain more stable across different decoder depths in multi-head configurations, especially for PTv3. These observations support our hypothesis that multi-head decoders capture richer semantic representations. Since EMD is particularly sensitive to global point distribution and uniformity, the structural gains enabled by multi-head designs translate into improved EMD performance.

\textbf{Hausdorff Distance reveals outlier sensitivity at higher depths: }Our results show that Hausdorff Distance (HD) improves with decoder depth across all models. However, beyond a certain threshold (4-5 layers), we observe a decline in HD performance. As HD is particularly sensitive to outliers, this drop in performance can be attributed to deeper decoders generating outliers or scattered points in low-density regions. Notably, this degradation is significantly mitigated in multi-head models, especially for ModelNet40 and PTv3 on ShapeNetPart. By having each decoder head produce a smaller subset of points, the model implicitly encourages each point to carry higher semantic value, reducing redundancy and minimizing the risk of generating outliers in sparse areas.

\section{Conclusion}
In this paper, we present a novel multi-head design for point cloud reconstruction. Our approach introduces multiple parallel decoder heads that share a common latent vector, with each head reconstructing the full shape from a subset of points. Through extensive experiments across three backbone models, five decoder depths, and two datasets, we demonstrate that our multi-head architecture consistently outperforms traditional, single-head decoders. The proposed method is easily adaptable to existing encoder-decoder frameworks and offers richer semantic coverage without significantly increasing computational cost.

\textbf{Limitations. }

\textbullet~ \textbf{Performance vs training setup: } Our conclusions regarding metric trends may depend on specific training setups. For instance, introducing batch normalization to Light-AE on ModelNet40 alters the observed trends. Nonetheless, in all cases, multi-head decoders consistently outperform single-head counterparts, supporting our main findings. We will investigate further in future work.

\textbullet~ \textbf{Number of heads vs number of points: } In our experiments, we use two decoder heads. However, the optimal number of heads likely correlates with the number of points per shape. Models with more than two heads may perform better when reconstructing denser point clouds, where redundancy is higher. We leave further exploration to future work.

\bibliographystyle{unsrt}
\bibliography{references}

\begin{thebibliography}{10}

\bibitem{Qi_2017_CVPR}
Charles~R. Qi, Hao Su, Kaichun Mo, and Leonidas~J. Guibas.
\newblock Pointnet: Deep learning on point sets for 3d classification and segmentation.
\newblock In {\em Proceedings of the IEEE Conference on Computer Vision and Pattern Recognition (CVPR)}, July 2017.

\bibitem{NIPS2017_d8bf84be}
Charles~Ruizhongtai Qi, Li~Yi, Hao Su, and Leonidas~J Guibas.
\newblock Pointnet++: Deep hierarchical feature learning on point sets in a metric space.
\newblock In I.~Guyon, U.~Von Luxburg, S.~Bengio, H.~Wallach, R.~Fergus, S.~Vishwanathan, and R.~Garnett, editors, {\em Advances in Neural Information Processing Systems}, volume~30. Curran Associates, Inc., 2017.

\bibitem{Yang_2018_CVPR}
Yaoqing Yang, Chen Feng, Yiru Shen, and Dong Tian.
\newblock Foldingnet: Point cloud auto-encoder via deep grid deformation.
\newblock In {\em Proceedings of the IEEE Conference on Computer Vision and Pattern Recognition (CVPR)}, June 2018.

\bibitem{Wu_2015_CVPR}
Zhirong Wu, Shuran Song, Aditya Khosla, Fisher Yu, Linguang Zhang, Xiaoou Tang, and Jianxiong Xiao.
\newblock 3d shapenets: A deep representation for volumetric shapes.
\newblock In {\em Proceedings of the IEEE Conference on Computer Vision and Pattern Recognition (CVPR)}, June 2015.

\bibitem{10.1145/2980179.2980238}
Li~Yi, Vladimir~G. Kim, Duygu Ceylan, I-Chao Shen, Mengyan Yan, Hao Su, Cewu Lu, Qixing Huang, Alla Sheffer, and Leonidas Guibas.
\newblock A scalable active framework for region annotation in 3d shape collections.
\newblock {\em ACM Trans. Graph.}, 35(6), December 2016.

\bibitem{vakalopoulou:hal-01958236}
Maria Vakalopoulou, Guillaume Chassagnon, Norbert Bus, Rafael Marini~Silva, Evangelia~I. Zacharaki, Marie-Pierre Revel, and Nikos Paragios.
\newblock {AtlasNet: Multi-atlas Non-linear Deep Networks for Medical Image Segmentation}.
\newblock In {\em {International Conference on Medical Image Computing and Computer-Assisted Intervention}}, Granada, Spain, September 2018.

\bibitem{yuan2018pcn}
Wentao Yuan, Tejas Khot, David Held, Christoph Mertz, and Martial Hebert.
\newblock Pcn: Point completion network.
\newblock In {\em 2018 International Conference on 3D Vision (3DV)}, pages 728--737, 2018.

\bibitem{Tchapmi_2019_CVPR}
Lyne~P. Tchapmi, Vineet Kosaraju, Hamid Rezatofighi, Ian Reid, and Silvio Savarese.
\newblock Topnet: Structural point cloud decoder.
\newblock In {\em Proceedings of the IEEE/CVF Conference on Computer Vision and Pattern Recognition (CVPR)}, June 2019.

\bibitem{liu2020morphing}
Minghua Liu, Lu~Sheng, Sheng Yang, Jing Shao, and Shi-Min Hu.
\newblock Morphing and sampling network for dense point cloud completion.
\newblock In {\em Proceedings of the AAAI conference on artificial intelligence}, volume~34, pages 11596--11603, 2020.

\bibitem{NIPS2016_44f683a8}
Jiajun Wu, Chengkai Zhang, Tianfan Xue, Bill Freeman, and Josh Tenenbaum.
\newblock Learning a probabilistic latent space of object shapes via 3d generative-adversarial modeling.
\newblock In D.~Lee, M.~Sugiyama, U.~Luxburg, I.~Guyon, and R.~Garnett, editors, {\em Advances in Neural Information Processing Systems}, volume~29. Curran Associates, Inc., 2016.

\bibitem{li2018pointcloudgan}
Chun-Liang Li, Manzil Zaheer, Yang Zhang, Barnabas Poczos, and Ruslan Salakhutdinov.
\newblock Point cloud gan, 2018.

\bibitem{Xiang_2021_ICCV}
Peng Xiang, Xin Wen, Yu-Shen Liu, Yan-Pei Cao, Pengfei Wan, Wen Zheng, and Zhizhong Han.
\newblock Snowflakenet: Point cloud completion by snowflake point deconvolution with skip-transformer.
\newblock In {\em Proceedings of the IEEE/CVF International Conference on Computer Vision (ICCV)}, pages 5499--5509, October 2021.

\bibitem{yu2022point}
Xumin Yu, Lulu Tang, Yongming Rao, Tiejun Huang, Jie Zhou, and Jiwen Lu.
\newblock Point-bert: Pre-training 3d point cloud transformers with masked point modeling.
\newblock In {\em Proceedings of the IEEE/CVF conference on computer vision and pattern recognition}, pages 19313--19322, 2022.

\bibitem{Yan_2022_CVPR}
Xingguang Yan, Liqiang Lin, Niloy~J. Mitra, Dani Lischinski, Daniel Cohen-Or, and Hui Huang.
\newblock Shapeformer: Transformer-based shape completion via sparse representation.
\newblock In {\em Proceedings of the IEEE/CVF Conference on Computer Vision and Pattern Recognition (CVPR)}, pages 6239--6249, June 2022.

\bibitem{Yu_2021_ICCV}
Xumin Yu, Yongming Rao, Ziyi Wang, Zuyan Liu, Jiwen Lu, and Jie Zhou.
\newblock Pointr: Diverse point cloud completion with geometry-aware transformers.
\newblock In {\em Proceedings of the IEEE/CVF International Conference on Computer Vision (ICCV)}, pages 12498--12507, October 2021.

\bibitem{nichol2022point}
Alex Nichol, Heewoo Jun, Prafulla Dhariwal, Pamela Mishkin, and Mark Chen.
\newblock Point-e: A system for generating 3d point clouds from complex prompts.
\newblock {\em arXiv preprint arXiv:2212.08751}, 2022.

\bibitem{Luo_2021_CVPR}
Shitong Luo and Wei Hu.
\newblock Diffusion probabilistic models for 3d point cloud generation.
\newblock In {\em Proceedings of the IEEE/CVF Conference on Computer Vision and Pattern Recognition (CVPR)}, pages 2837--2845, June 2021.

\bibitem{Shu_2019_ICCV}
Dong~Wook Shu, Sung~Woo Park, and Junseok Kwon.
\newblock 3d point cloud generative adversarial network based on tree structured graph convolutions.
\newblock In {\em Proceedings of the IEEE/CVF International Conference on Computer Vision (ICCV)}, October 2019.

\bibitem{xie2020grnet}
Haozhe Xie, Hongxun Yao, Shangchen Zhou, Jiageng Mao, Shengping Zhang, and Wenxiu Sun.
\newblock Grnet: Gridding residual network for dense point cloud completion.
\newblock In {\em European conference on computer vision}, pages 365--381. Springer, 2020.

\bibitem{wu2024point}
Xiaoyang Wu, Li~Jiang, Peng-Shuai Wang, Zhijian Liu, Xihui Liu, Yu~Qiao, Wanli Ouyang, Tong He, and Hengshuang Zhao.
\newblock Point transformer v3: Simpler faster stronger.
\newblock In {\em Proceedings of the IEEE/CVF Conference on Computer Vision and Pattern Recognition}, pages 4840--4851, 2024.

\bibitem{chang2015shapenetinformationrich3dmodel}
Angel~X. Chang, Thomas Funkhouser, Leonidas Guibas, Pat Hanrahan, Qixing Huang, Zimo Li, Silvio Savarese, Manolis Savva, Shuran Song, Hao Su, Jianxiong Xiao, Li~Yi, and Fisher Yu.
\newblock Shapenet: An information-rich 3d model repository, 2015.

\bibitem{kingma2014adam}
Diederik~P Kingma.
\newblock Adam: A method for stochastic optimization.
\newblock {\em arXiv preprint arXiv:1412.6980}, 2014.

\end{thebibliography}

\end{document}